\documentclass[pdflatex,sn-mathphys-num,iicol]{sn-jnl}

\usepackage{graphicx}%
\usepackage{multirow}%
\usepackage{amsmath,amssymb,amsfonts}%
\usepackage{amsthm}%
\usepackage{mathrsfs}%
\usepackage[title]{appendix}%
\usepackage{xcolor}%
\usepackage{textcomp}%
\usepackage{manyfoot}%
\usepackage{booktabs}%
\usepackage{algorithm}%
\usepackage{algorithmicx}%
\usepackage{algpseudocode}%
\usepackage{listings}%
\usepackage{makecell}

\theoremstyle{thmstyleone}%
\theoremstyle{thmstyletwo}%

\theoremstyle{thmstylethree}%

\begin{document}

\title[Article Title]{Vision-Language Models for Deployable Social Robot Navigation: Bridging Semantic Reasoning and Low-Level Control}


\author[1]{\fnm{Runji} \sur{Cai}}\email{junnkidesu@gmail.com}
\author[2]{\fnm{Toshihiko} \sur{Yamasaki}}\email{yamasaki@cvm.t.u-tokyo.ac.jp}
\author*[1]{\fnm{Ling} \sur{Xiao}}\email{ling@ist.hokudai.ac.jp}

\affil[1]{\orgdiv{Graduate School of Information Science and Technology}, \orgname{Hokkaido University}, \city{Sapporo}, \postcode{060-0814}, \country{Japan}}

\affil[2]{\orgdiv{Graduate School of Information Science and Technology}, \orgname{The University of Tokyo}, \city{Tokyo}, \postcode{113-8656}, \country{Japan}}


\abstract{Social robot navigation (SRN) requires more than geometric path planning; it demands understanding human intentions, social norms, and contextual cues to generate socially compliant behaviors. Although classical navigation methods provide reliable metric planning and collision avoidance, they often lack the semantic reasoning capabilities necessary for operation in complex human-centered environments.
Recent advances in Vision-Language Models (VLMs) have opened new opportunities for SRN by enabling high-level VLM understanding, commonsense reasoning, and natural language interaction. However, a fundamental challenge remains: how to integrate VLMs into real-time, safety-critical navigation systems and reliably translate their high-level reasoning into grounded navigation actions.
In this survey, we present a unified perspective of VLM-based SRN and organize existing approaches into three interconnected components: high-level VLM reasoning, low-level planning and control, and intermediate mechanisms that bridge reasoning and action. Based on this perspective, we propose a structured roadmap for coupling VLMs with navigation systems, covering semantic reasoning, evaluators, spatial grounding, intermediate representations, and control modules. The roadmap highlights both the strengths of VLMs and the necessity of hybrid architectures for practical deployment.
We further review representative datasets and evaluation platforms developed for SRN. Finally, we discuss key open challenges. This survey aims to provide a foundation for building reliable, socially compliant, and deployable VLM-enabled navigation systems.}


\maketitle

\section{Introduction}
\label{sec:intro}

Rapid urbanization, aging populations, and increasing demand for automation are accelerating the deployment of robots in human-centered environments~\cite{singamaneni2024survey}. Unlike industrial robots in structured settings, robots deployed in social environments must navigate shared spaces such as healthcare facilities, educational settings, and public sidewalks, where they frequently interact with nearby humans~\cite{mavrogiannis2023core, han2024human, payandeh2025social}.

In such environments, navigation is no longer a purely geometric problem. Beyond reaching a destination safely and efficiently, robots must generate motion that is smooth, legible, and socially acceptable to nearby people~\cite{kruse2013human, mirsky2024conflict}. In Human-Robot Interaction (HRI), motion itself serves as a form of communication: humans infer robot intent from trajectories, and legible motion that reveals the robot’s goal differs fundamentally from merely predictable motion~\cite{dragan2013legibility}. Moreover, SRN requires respecting proxemic norms and context-dependent social conventions, rather than modeling humans merely as moving obstacles~\cite{rios2015proxemics}.

This shift fundamentally changes how navigation systems should be evaluated. A robot may reach its destination efficiently, yet still fail in real-world scenarios if its behavior is socially inappropriate, for example, by cutting in front of a person or intruding into personal space~\cite{mavrogiannis2023core, francis2025principles}. Empirical studies also show that robot navigation strategies directly influence human behavior~\cite{mavrogiannis2019effects}. As a result, success must be measured not only by navigation efficiency, but also by human-centered outcomes such as comfort, perceived safety, legibility, and acceptance.

\begin{figure}[t]
\centering
\includegraphics[width=\linewidth]{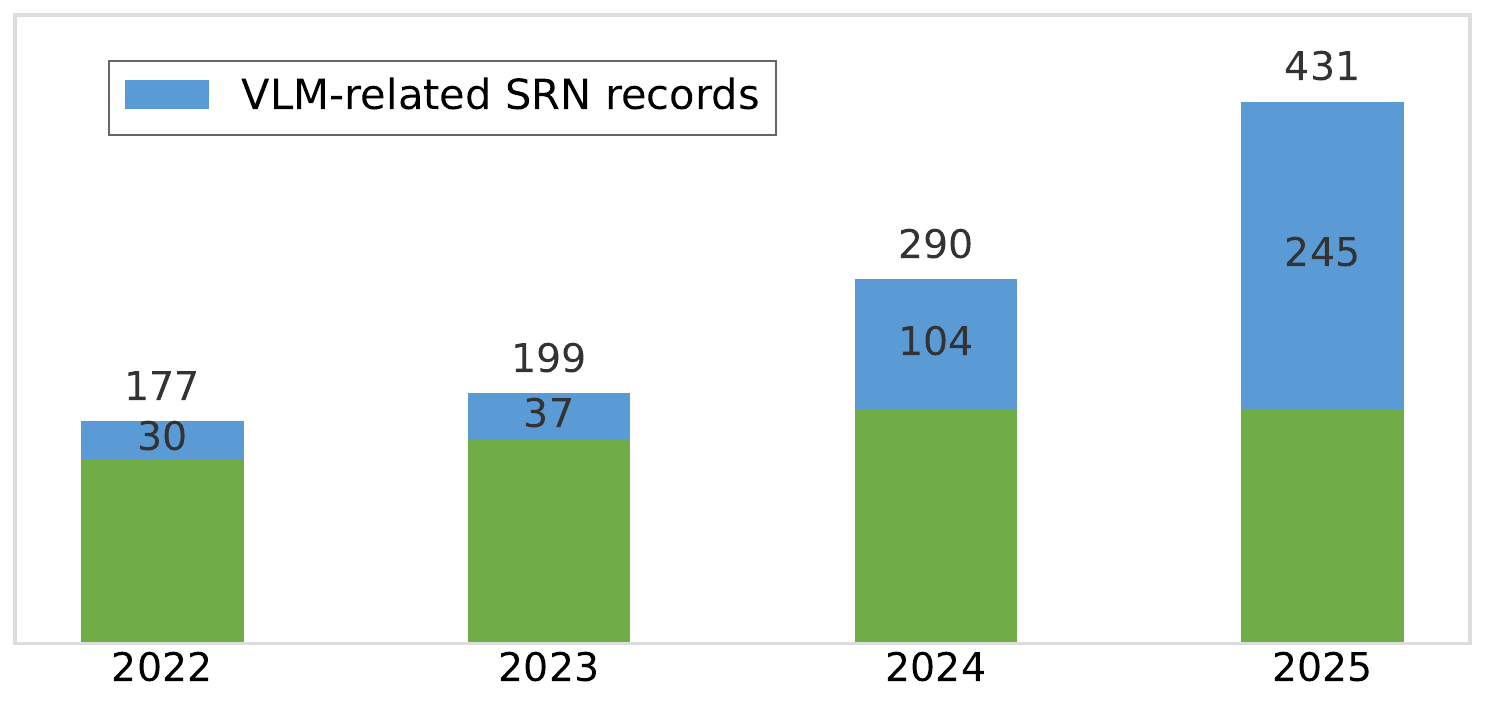}
\caption{The counts represent the raw Google Scholar search results for publications related to SRN and VLM-based SRN from 2022 to 2025. This graph illustrates the growing popularity of this research topic over time.}
\label{fig:srn_vlm_records}
\end{figure}

From a system perspective, achieving such behavior requires integrating three components: high-level reasoning for semantic understanding and social decision-making, low-level planning and control for generating feasible and safe trajectories, and a bridging mechanism that translates high-level reasoning into executable actions~\cite{xiao2022motion, firoozi2025foundation}. While recent advances in vision-language models (VLMs), often supported by large language models (LLMs), have significantly improved high-level reasoning capabilities~\cite{achiam2023gpt, bommasani2021opportunities}, and classical planning and learning-based controllers provide effective low-level planning and control~\cite{xiao2022motion, kretzschmar2016socially}, the connection between these two levels remains underexplored. This creates a barrier between semantic reasoning and socially grounded motion execution.

This gap is also reflected in existing surveys. Social navigation surveys~\cite{kruse2013human, singamaneni2024survey} and VLM-oriented surveys~\cite{han2025multimodal, zhang2024vision} tend to address socially aware motion generation and vision-language reasoning in isolation. As a result, few surveys systematically examine the \emph{bridging interface} that connects high-level VLM reasoning to deployable, socially compliant mobile-robot motion, as shown in Table~\ref{tab:survey_comparison}. In addition, recent benchmarking efforts further highlight the lack of standardized evaluation protocols for assessing deployment-oriented performance~\cite{chhetri2025short, francis2023benchmarking}. This limitation slows progress and limits the extent to which VLMs can improve social navigation, despite the rapid growth of VLM-enabled SRN research, as shown in Fig.~\ref{fig:srn_vlm_records}.

This survey addresses this gap by focusing on how high-level VLM reasoning can be grounded into real-time, safe, and socially compliant navigation. By organizing existing methods from this perspective, we provide a structured roadmap toward building reliable and deployable SRN systems. Specifically, our contributions are threefold:

\begin{table*}[t]
\centering
\footnotesize
\setlength{\tabcolsep}{2pt}
\renewcommand{\arraystretch}{1.08}
\caption{Positioning of this survey among previous surveys.}
\label{tab:survey_comparison}
\begin{tabular}{@{}>{\raggedright\arraybackslash}m{0.24\textwidth}
                >{\raggedright\arraybackslash}m{0.06\textwidth}
                >{\raggedright\arraybackslash}m{0.25\textwidth}
                >{\raggedright\arraybackslash}m{0.41\textwidth}@{}}
\toprule
Previous Surveys & Year & Foundation Method / Model Coverage & Main Focus \\
\midrule
Kruse, Thibault, et al.~\cite{kruse2013human} 
& 2013 
& Classical planning methods; ML for robotic motion/navigation 
& Classification of human-aware navigation methods for wheeled robots \\ \midrule

Rios-Martinez, Jorge, et al.~\cite{rios2015proxemics} 
& 2015 
& None 
& Social conventions, proxemics, and context-aware spaces for socially-aware navigation \\ \midrule

Chik, S. F., et al.~\cite{chik2016review} 
& 2016 
& Classical planning methods and human-motion prediction models 
& Social-aware navigation frameworks for service robots in dynamic human environments, focusing on navigation components, prediction models, and robot platforms \\ \midrule

Charalampous, Konstantinos, et al.~\cite{charalampous2017recent} 
& 2017 
& Classical ML/probabilistic models for robotic motion/navigation 
& Robot perception, social-awareness methods, datasets, and open challenges in SRN \\ \midrule

Cheng, Jiyu, et al.~\cite{cheng2018autonomous} 
& 2018 
& Classical planning methods; ML/RL for robotic motion/navigation 
& Autonomous mobile robot navigation in human environments, including dynamic humans, HRI, group information, navigation strategies, and evaluation metrics \\ \midrule

Lambert, Alexis, et al.~\cite{lambert2020systematic} 
& 2020 
& None 
& Systematic review of social robotics research topics and applications \\ \midrule

M{\"o}ller, Ronja, et al.~\cite{moller2021survey} 
& 2021 
& Embodied AI for perception; ML for human-aware navigation 
& Interdisciplinary survey integrating active vision, visual navigation, HRI, human behavior modeling, activity recognition, trajectory forecasting, and simulation-based evaluation \\ \midrule 
 
Wang, Junxian, et al.~\cite{wang2022metrics} 
& 2022 
& RL for robotic motion/navigation 
& Metrics and evaluation protocol for socially conforming crowd navigation \\ \midrule

Mavrogiannis, Christoforos, et al.~\cite{mavrogiannis2023core} 
& 2023 
& None 
& Planning, behavior design, interaction modeling, and evaluation challenges in SRN \\ \midrule

Mirsky, Reuth, et al.~\cite{mirsky2024conflict} 
& 2024 
& ML for robotic motion/navigation 
& Conflict avoidance taxonomy and open problems in social navigation \\ \midrule

Singamaneni, Phani Teja, et al.~\cite{singamaneni2024survey} 
& 2024 
& ML for robotic motion/navigation 
& Taxonomy of socially aware robot navigation across robot types, planning, situation awareness, and evaluation \\ \midrule

Francis, Anthony, et al.~\cite{francis2025principles} 
& 2025 
& ML for robotic motion/navigation
& Principles, guidelines, and evaluation framework for SRN across metrics, scenarios, benchmarks, datasets, and simulators \\ \midrule

Ours 
& 2026 
& VLM for SRN
& A working definition of VLM-based social robot
navigation, a structured roadmap for coupling VLMs with
social navigation systems, and a comprehensive analysis of datasets, evaluation platforms, and open challenges. \\
\bottomrule
\end{tabular}
\end{table*}

\begin{itemize}

\item \textbf{A working definition of VLM-based SRN.} 
We define the VLM-based formulation of SRN as translating high-level VLM reasoning into executable, safe and socially compliant robot behavior, and define the roles of high-level reasoning, low-level planning and control, and the bridging function (Section~\ref{sec:definition}).
\item \textbf{A structured roadmap for coupling VLMs with social robot navigation systems.} 
We propose a taxonomy of five paradigms organized by the interfaces that connect VLM reasoning to socially compliant robot navigation systems, explicitly annotating each paradigm with its level of supporting evidence (Section~\ref{sec:taxonomy}).
\item \textbf{A comprehensive analysis of datasets, evaluation platforms, and open challenges.} 
We consolidate datasets and evaluation platforms, and identify the key challenges that limit the transition from VLM-based reasoning to real-world deployment performance (Sections~\ref{sec:datasets}--\ref{sec:challenges_future}).

\end{itemize}

\section{A System-Level Formulation of VLM-Based SRN}
\label{sec:definition}

In this survey, VLM-based SRN can be abstracted as a hierarchical formulation that links
high-level VLM reasoning to low-level planning and control through an explicit
bridging representation. The high-level module interprets visual observations and language instructions. The bridging function
converts these outputs into goals, reference behaviors, cost weights, and
constraints that can be directly used by the low-level planner to optimize feasible trajectories in real time.
This definition keeps VLM reasoning separate from direct actuation:
VLMs provide semantic and social structure, while the controller remains
responsible for dynamic feasibility, safety, and execution.

\subsection{High-Level VLM Reasoning}
\label{sec:def-highlevel}

In VLM-based SRN, the high-level module converts visual observations and language instructions into high-level interface variables that can guide downstream planning. Let $I^t$ denote the robot-view RGB image, and let $\rho^t$ denote the textual prompt or language instruction. We summarize the possible forms of VLM outputs as
\begin{equation}
    B_h^t=
    \Phi_\theta(I^t,\rho^t),
    \label{eq:highlevel-vlm-interface}
\end{equation}
where $\Phi_\theta$ is a VLM or VLA model that maps visual
observations and language instructions into a semantic high-level signal.
\begin{equation}
    B_h^t =
    (r_t,q_t,M_t,z_t,a_t^h)
    \in \mathcal B_h,
    \label{eq:highlevel-vlm-interface-output}
\end{equation}

The set $\mathcal B_h$ should be
read as a general interface space rather than a set of mandatory joint outputs:
a particular method typically instantiates one or a subset of these output
forms. For notational convenience, components that are not instantiated by a particular
method are treated as absent or null entries in Eq.~\eqref{eq:highlevel-vlm-interface-output}. Specifically, $r_t$ denotes language reasoning or social-norm
explanation, $q_t$ denotes a score or selector over candidate goals, paths, or
actions, $M_t$ denotes a semantic or language-queryable map, $z_t$
denotes an intermediate representation such as waypoints or attention maps, and
$a_t^h$ denotes a high-level or mid-level action. Existing systems instantiate
different parts of this interface, including VLM-based scoring or reward modeling~\cite{rocamonde2023vision}, waypoint prediction~\cite{shah2023vint}, 
semantic map querying~\cite{chen2023open},
and VLA action prediction~\cite{cheng2024navila,hirose2025omnivla}.

\subsection{Bridging Definition}
\label{sec:def-bridging}

The bridging module translates the high-level VLM signal into quantities that can be directly used by the planner or controller. We define this interface as
\begin{equation}
\begin{aligned}
    \eta_t
    &=
    \Gamma_\omega
    \left(
        B_h^{t},
        o_t
    \right) \\
    &=
    \left(
        g_t,\psi_t,u_t^h,\lambda_t,\mathcal K_t
    \right),
\end{aligned}
\label{eq:bridging-connected}
\end{equation}
where $\Gamma_\omega$ denotes the bridging module parameterized by $\omega$.
$B_h^t$ denotes the most recently available high-level VLM output at time $t$, and
$o_t$ denotes the robot-side observation available to the bridge and the planner,
which may include $I^t$, the robot state, and environment information.
The components of $\eta_t$ specify how high-level semantics are converted into planner-facing quantities.
Here $g_t$ is a semantic subgoal. $\psi_t$ denotes a compact social-context descriptor that may encode human occupancy, predicted human-motion patterns, interaction relations or graphs, passing preferences, and social norms~\cite{kretzschmar2016socially,alahi2016social,mohamed2020social}. 
When the high-level output contains an action component $a_t^h$, the bridge
may convert it into a planner-facing reference behavior $u_t^h$.
$\lambda_t$ denotes adaptive cost weights, and $\mathcal K_t$ denotes hard planning constraints.
In this view, the bridge is the formal interface through which VLM
semantics become controller-level costs, constraints, or reference behaviors, as
in VLM-integrated navigation interfaces~\cite{cheng2024navila}.

The bridge also resolves the temporal mismatch between high-level reasoning and low-level planning. VLM inference typically updates at a lower rate than the
motion controller, often due to latencies ranging from hundreds of milliseconds to several seconds~\cite{vasu2025fastvlm}. In contrast, deployable local navigation and
socially compatible motion control require frequent replanning and control
updates, commonly around or above 10~Hz~\cite{li2025integrated,trautman2015robot}.
Thus, $\Gamma_\omega$ can be interpreted as an asynchronous interface: it refreshes
$\eta_t$ when new VLM outputs or human-motion predictions are available, while
the low-level controller reuses the latest $\eta_t$ over faster control cycles.

\subsection{Low-Level Planning and Control}
\label{sec:def-lowlevel}

Given the planner-facing representation $\eta_t$, the low-level planner
optimizes an executable trajectory
$\tau_t=(x_{t:t+K},u_{t:t+K-1})$, where $x_{t:t+K}=(x_t,x_{t+1},\ldots,x_{t+K})$ denotes the state sequence and
$u_{t:t+K-1}=(u_t,u_{t+1},\ldots,u_{t+K-1})$ denotes the control sequence, and
$K$ is the planning horizon:
\begin{equation}
\tau_t^\star
=
\arg\min_{\tau_t\in\mathcal{F}_t(\eta_t;o_t)}
C_{\eta_t}(\tau_t;o_t),
\label{eq:def-lowlevel}
\end{equation}
where $C_{\eta_t}(\tau_t;o_t)$ is the navigation cost conditioned on $\eta_t$ and
the current observation $o_t$, and $\mathcal F_t(\eta_t;o_t)$ is the feasible
trajectory set.

The feasible set summarizes collision avoidance, social clearance, and hard
constraints:
\begin{equation}
\begin{aligned}
\mathcal{F}_t(\eta_t;o_t)
=
\Bigl\{
\tau_t \Bigm|
&x_t=\hat{x}_t,\\
&x_{k+1}=F(x_k,u_k),\\
&x_k\in\mathcal{X}_{\mathrm{free}},\\
&d(x_k,\mathcal{O}_t)\ge d_{\mathrm{obs}},\\
&d(x_k,\hat h_k^{\,i})\ge d_{\min},\\
&\tau_t \models \mathcal K_t,\\
&\text{for all admissible } k,\ \forall i\in\mathcal I_t
\Bigr\}.
\end{aligned}
\label{eq:def-feasible-set}
\end{equation}
Here, $\hat{x}_t$ denotes the current robot state estimated from the observation
$o_t$, and the constraint $x_t=\hat{x}_t$ ensures that the planned trajectory
starts from the current robot state. $\mathcal{X}_{\mathrm{free}}$ denotes the
collision-free state space, and $F$ denotes the robot dynamics model. The function
$d(\cdot,\cdot)$ denotes a distance metric used to evaluate obstacle and
human-clearance constraints. The admissible ranges of $k$ are determined by the
corresponding state and control sequences. Specifically, the dynamics constraint is
defined over the control horizon, whereas the state, obstacle-clearance, and
social-clearance constraints are defined over the state prediction horizon.
$\mathcal{O}_t$ denotes the obstacle set, $d_{\mathrm{obs}}$ is the minimum
obstacle clearance, $d_{\min}$ is the minimum social safety distance, and
$\mathcal I_t$ denotes the set of nearby humans considered at time $t$. The term
$\hat h_k^{\,i}$ denotes the predicted state of human $i$ at future time step $k$,
obtained from the predicted human trajectory set contained in or derived from $o_t$. The relation $\tau_t \models \mathcal K_t$ means that the planned trajectory
$\tau_t$ satisfies all bridge-provided hard constraints in $\mathcal K_t$.

Therefore, the constraint
$d(x_k,\hat h_k^{\,i})\ge d_{\min}$ enforces socially safe clearance
from each nearby human over the planning horizon
\cite{alahi2016social,mohamed2020social}.

\subsection{Closed-Loop Receding-Horizon Execution}
\label{sec:def-closed-loop}

Combining high-level VLM reasoning, bridging, and low-level planning and control yields a
closed-loop VLM-based optimization formulation for SRN. 
The resulting receding-horizon execution is therefore:
\begin{equation}
\begin{aligned}
    u_t^\star
    &=
    \operatorname{first}_u(\tau_t^\star).
\end{aligned}
\label{eq:connected-closed-loop-optimization}
\end{equation}
Here, $\operatorname{first}_u(\cdot)$ extracts the first control input from the
optimized control sequence. The optimized control action $u_t^\star$ is executed, producing the next observation. The next observation is then used to update perception, human-motion prediction, high-level VLM reasoning and the bridge.
The planner cost can be written compactly as
\begin{equation}
\begin{aligned}
C_{\eta_t}(\tau_t;o_t)
=
&\sum_{k=t}^{t+K-1}\\
\Big[
&\lambda_{g,t} C_{goal}(x_k,g_t)\\
&+\lambda_{o,t} C_{obstacle}(x_k,\mathcal O_t)\\
&+\lambda_{s,t} C_{social}
(x_k,\psi_t)\Big]\\
&+\lambda_{v,t} C_{vlm}(\tau_t,u_t^h)
\\
&+\lambda_{l,t} C_{legibility}(\tau_t,g_t)\\
&+\lambda_{n,t} C_{naturalness}(\tau_t,\psi_t),
\end{aligned}
\label{eq:def-cost}
\end{equation}
where the cost weights are

\begin{equation}
\lambda_t=
(\lambda_{g,t},\lambda_{o,t},\lambda_{s,t},
\lambda_{v,t},\lambda_{l,t},\lambda_{n,t}),
\,
\lambda_t \in \mathbb{R}_{\geq 0}^{6}.
\label{eq:def-weights}
\end{equation}
Here, $\lambda_{g,t}$, $\lambda_{o,t}$, $\lambda_{s,t}$,
$\lambda_{v,t}$, $\lambda_{l,t}$, and $\lambda_{n,t}$ weight the goal,
obstacle, social, VLM-derived reference, legibility, and naturalness costs,
respectively. $C_{goal}$ measures progress toward the bridge-provided goal or subgoal
$g_t$~\cite{anderson2018evaluation}. $C_{obstacle}$ penalizes physical collision
risk, while $C_{social}$ penalizes socially inappropriate motion according to
the social-context descriptor $\psi_t$~\cite{luo2025goal,wang2022metrics}. $C_{vlm}$ captures consistency between the planned trajectory and the
bridge-provided VLM-derived reference behavior $u_t^h$~\cite{song2024vlm}.
When the high-level output does not contain an action component, the entire term
$\lambda_{v,t} C_{vlm}(\tau_t,u_t^h)$ is set to zero, and $u_t^h$ is treated as
inactive. Finally, $C_{legibility}$ encourages
legible intent expression~\cite{dragan2013legibility}, and $C_{naturalness}$
encourages smooth and socially natural motion~\cite{wang2022metrics,kretzschmar2016socially}. This formulation unifies
semantic reasoning, human-motion prediction, bridging, and feasible control
within a single planner interface.

\section{Coupling VLMs with SRN Systems}
\label{sec:taxonomy}

Classical methods form the foundation of robot navigation. They typically represent the environment using occupancy grids, cost maps, or hand-crafted geometric features, leaving the rich semantic and social context largely unmodeled. Global planners~\cite{lavalle2001randomized,kavraki2002probabilistic,loganathan2023systematic} search for collision-free routes in known or partially known environments, but they do not inherently reason about object semantics, human intent, or social norms. Local planners such as Dynamic Window Approach (DWA) and Timed Elastic Band (TEB)~\cite{fox2002dynamic,rosmann2012trajectory} support real-time obstacle avoidance and local trajectory generation, yet their decisions are typically short-horizon and geometry-driven. Social-force models~\cite{helbing1995social} introduced a behavior-level description of human motion by modeling pedestrians as being influenced by goals, other agents, and environmental constraints. This provides an early modeling basis for socially aware navigation. In contrast to these geometry and model driven approaches, recent vision-language foundation models further expand the semantic perception capability of navigation systems by enabling open-vocabulary visual understanding and semantic reasoning, thus supporting VLM-based perception for navigation~\cite{liu2023visual}.

\begin{figure*}
    \centering
    \includegraphics[width=\linewidth]{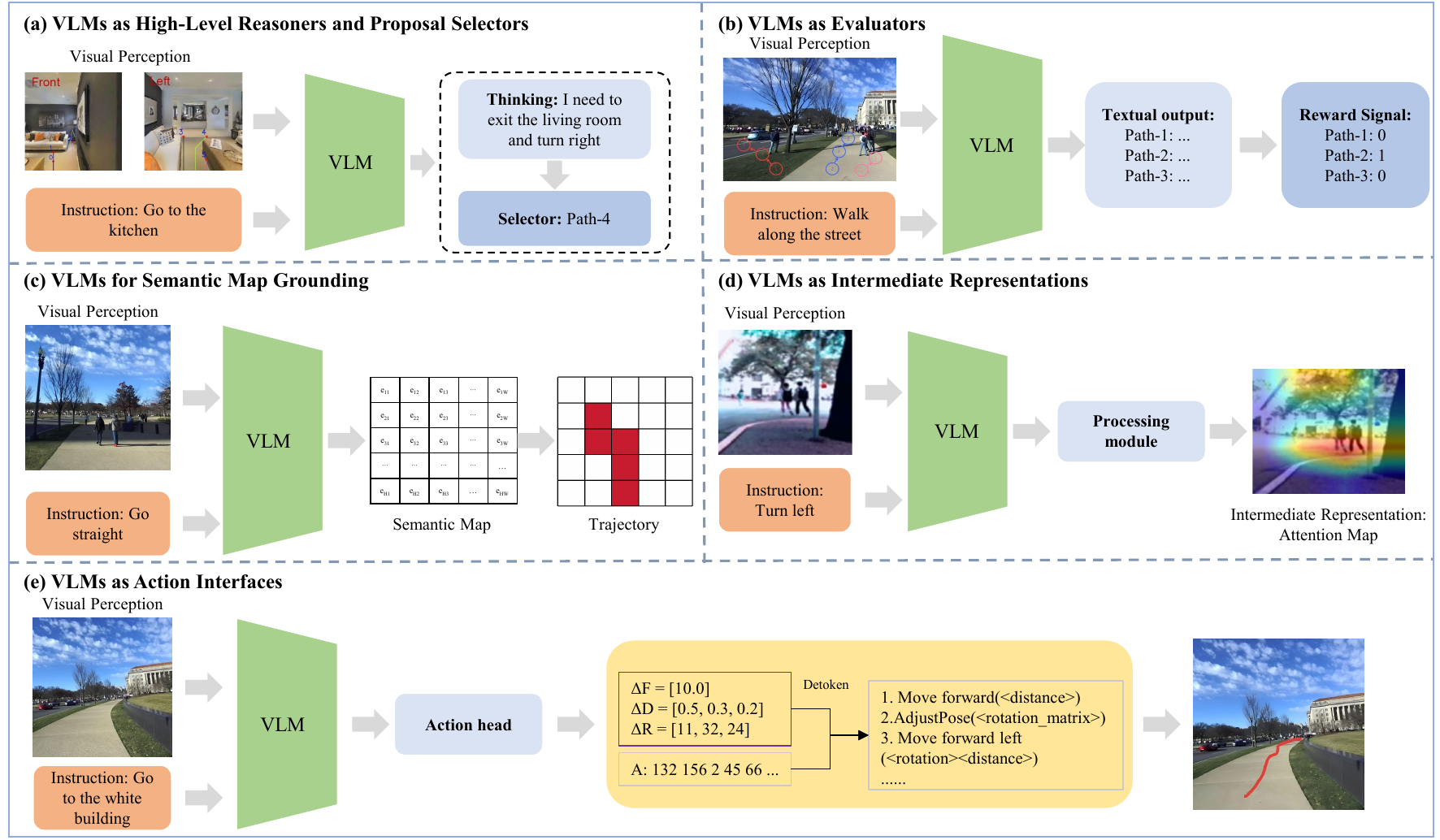}
    \caption{Five paradigms for integrating VLMs into SRN.}
	\label{fig:taxonomy}
\end{figure*}

\subsection{VLMs as High-Level Reasoners and Proposal Selectors}
Large VLMs provide a flexible interface for connecting visual observations with semantic reasoning, making them attractive as high-level decision modules in SRN. Early multimodal adaptation methods such as LLaMA-Adapter~\cite{zhang2023llama} show that frozen language models can be efficiently extended to image-conditioned instruction following and visual question answering (VQA), establishing the basic capability needed for language-grounded visual reasoning. In robotics, this capability is often used at a level above direct motor control. TPVQA~\cite{zhang2023grounding} converts preconditions and effects into VQA queries, allowing a VLM to verify action affordances, detect execution failures, and trigger replanning. Similarly, VLM-PC~\cite{chen2025commonsense} uses a VLM as a history-conditioned skill planner for legged locomotion, selecting high-level skills such as crawling, climbing, or detouring from egocentric images and past interaction history. These systems suggest a common design principle that VLMs are most reliable when asked to reason over semantic state, task progress, or skill choices.

A second line of work strengthens this role by improving the spatial grounding of VLM reasoning. General VLMs can recognize and describe visual scenes, but social navigation requires estimating how people, objects, and free space constrain safe and socially compliant motion. SpatialVLM~\cite{chen2024spatialvlm} addresses this spatial grounding limitation by training on large-scale spatial VQA data synthesized from real-world images, supporting metric estimates of distances, object sizes, and spatial relations from 2D inputs. SpatialRGPT~\cite{cheng2024spatialrgpt} further introduces region-aware and depth-aware spatial reasoning, allowing the model to reason about user-specified regions and local object relations rather than only image-level context. These advances are especially relevant for navigation pipelines that annotate images with candidate goals, regions, waypoints, or paths: the VLM must not only recognize scene semantics but also compare concrete spatial alternatives.

These requirements motivate a proposal-selection paradigm in which geometric or learned modules first generate feasible candidate goals, and the VLM then selects or scores the option most consistent with task semantics and social context, as illustrated in Fig.~\ref{fig:taxonomy} (a). LeLaN~\cite{hirose2024lelan} uses VLMs to annotate egocentric videos with object-navigation instructions and distills these annotations into a fast language-conditioned policy. Navi2Gaze~\cite{zhu2025navi2gaze} generates candidate action proposals, then uses VLM scoring to choose the action that best supports task-aware object gazing and subsequent manipulation. Mem2Ego~\cite{zhang2025mem2ego} extends proposal selection to long-horizon object navigation by retrieving global memory cues, projecting candidate markers into the egocentric view, and prompting the VLM to choose the next marker to explore.

In SRN, this proposal-selection pattern becomes particularly important because geometrically valid paths may still violate human expectations. Social-LLaVA and SNEI~\cite{payandeh2025social} show that social navigation can be formulated as language-grounded perception and prediction, followed by reasoning, action selection, and explanation generation, enabling a fine-tuned VLM to produce socially compliant high-level actions in public space scenarios. Building directly on the proposal-selection pattern, Fang et al.~\cite{fang2026obstacles} generate collision-free candidate paths from obstacle maps and predicted human motion, project these paths into the image, and use a fine-tuned VLM to select the socially preferred path for a downstream controller. Overall, these works indicate that VLM-based SRN is most promising when high-level social reasoning is coupled with explicit candidate generation, spatial grounding, and safety-aware low-level planning and control.

\subsection{VLMs as Evaluators}

Given high-level perceptual inputs, VLMs can serve as external evaluators for robot navigation by judging whether candidate behaviors are consistent with task goals and social context. They can also produce reward-like scores for downstream planning or optimization, as illustrated in Fig.~\ref{fig:taxonomy} (b). 

Following recent work on foundation models for reward design, Kwon et al.~\cite{kwon2023reward} shows that an LLM can act as a proxy reward function by evaluating whether agent behavior satisfies natural-language objectives. This idea naturally extends to vision-language settings, where visual observations are included in the evaluation input. Although not specific to SRN, VLM-RMs~\cite{rocamonde2023vision} and Language-Image Value (LIV)~\cite{ma2023liv} provide related examples of using pretrained vision-language representations for reward or value estimation. These works motivate the use of VLMs as evaluators rather than direct controllers.

\begin{figure}[t]
\centering
\includegraphics[width=\linewidth]{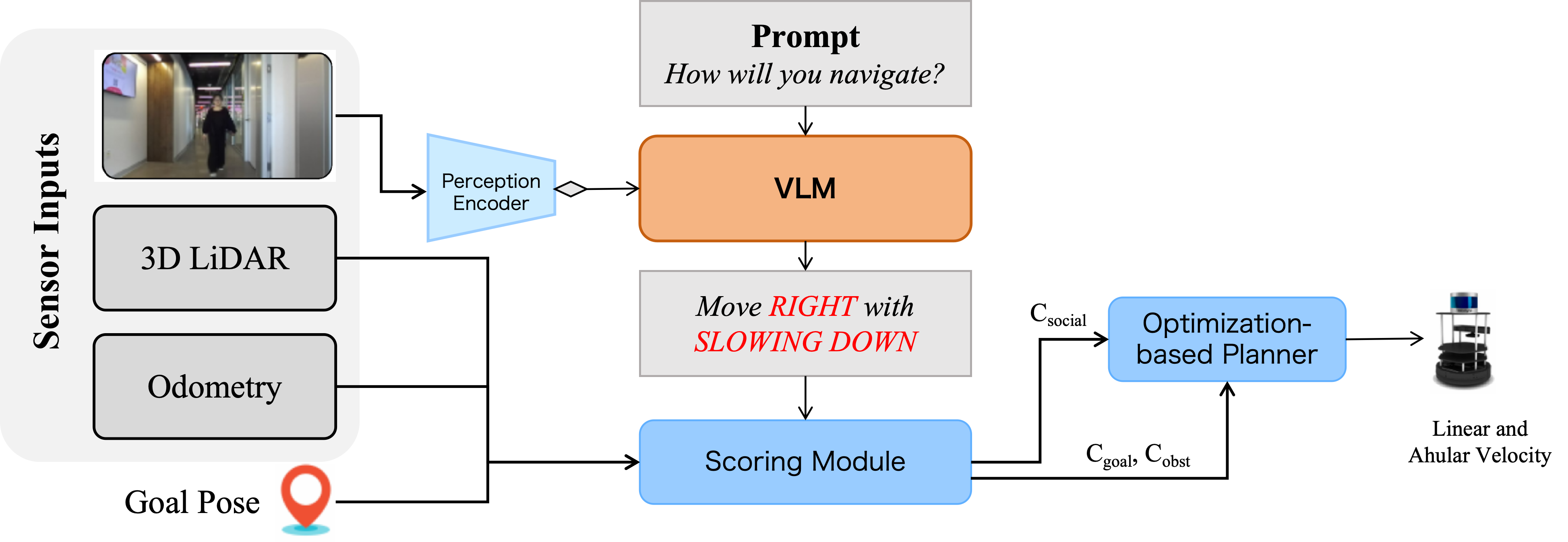}
\caption{VLM-Social-Nav adds a VLM-derived social score to a classical planner; adapted from~\cite{song2024vlm}.}
\label{fig:tax-scorer-vlmsocialnav}
\end{figure}

In navigation, this evaluator role often appears as candidate scoring. VLM-Social-Nav~\cite{song2024vlm}, as shown in Fig.~\ref{fig:tax-scorer-vlmsocialnav}, instantiates this evaluator role by converting VLM recommendations into a social cost for robot navigation. It prompts the VLM with visual observations and social navigation instructions, obtains a structured recommendation for socially appropriate motion, and converts this recommendation into a social cost term. This cost is added to the planner objective together with goal-reaching and obstacle-avoidance costs. Thus, the final action is still selected by the planner, but the planner is biased toward behavior that is more socially appropriate.

In general, these methods share a common integration pattern: the VLM evaluates and the low-level planning and control module executes. This division adds semantic and social reasoning to navigation without transferring low-level planning and control to the VLM. This separation can improve modularity and facilitate integration with existing navigation stacks.

\subsection{VLMs for Semantic Map Grounding}
\label{sec:paradigm-map}

Using VLMs as evaluators can improve action or path selection, but this paradigm provides only limited support for persistent or queryable scene understanding. Traditional robotic scene-understanding methods usually rely on labeled datasets to train supervised models for predefined tasks, which restricts their ability to handle open-vocabulary queries and unseen concepts. Although LLMs have enabled task planning from human instructions~\cite{chen2023open}, their application to real-world robotic tasks remains limited when they are not grounded in the surrounding scene. This motivates using vision-language representations as queryable semantic maps, as illustrated in Fig.~\ref{fig:taxonomy} (c).

Contrastive Language-Image Pretraining (CLIP) based scene representations~\cite{shafiullah2022clip} provide an early technical foundation for this shift. Cross-modal map learning~\cite{georgakis2022cross} predicts top-down semantic maps from language and egocentric observations, and then generates waypoint sequences toward the goal. Shah et al.~\cite{shah2023lm} uses a grounded language model to extract landmarks from language instructions, align them with nodes in a visual topological map via CLIP, and use the resulting landmark node associations for long-horizon robot navigation.

\begin{figure*}[t]
\centering
\includegraphics[width=\linewidth]{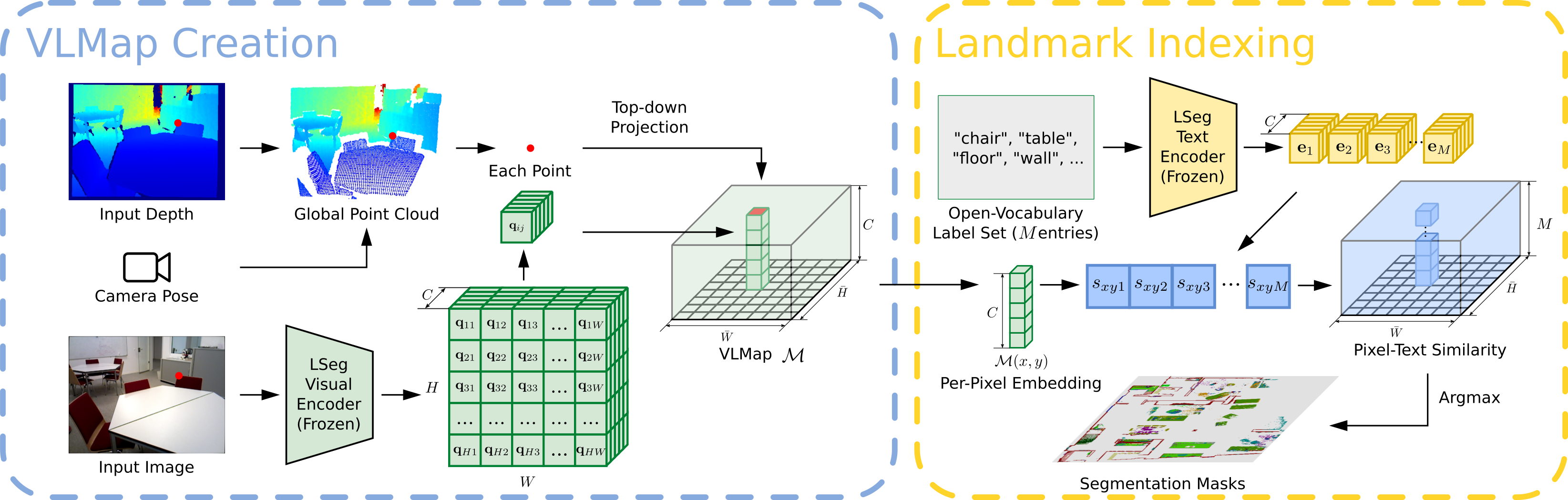}
\caption{VLMaps aligns vision-language features with a 3D map to support open-vocabulary spatial queries; adapted from~\cite{huang2023visual}.}
\label{fig:tax-map-vlmaps}
\end{figure*}

VLMaps~\cite{huang2023visual} extends queryable semantic mapping to robot navigation by fusing pretrained vision-language features with a 3D reconstruction of the physical world, as shown in Fig.~\ref{fig:tax-map-vlmaps}. This representation connects open-vocabulary language grounding with the spatial precision of geometric maps. As a result, language instructions can be translated into map-localized spatial goals, and navigation can use both geometric structure and semantic information. 

Recent work extends this line from vision-language mapping to richer forms of open-vocabulary scene understanding. OpenScene~\cite{peng2023openscene} predicts dense features for 3D scene points that are co-embedded with text and image pixels in CLIP feature space, enabling zero-shot queries over objects, materials, and other semantic categories. ConceptGraphs~\cite{gu2024conceptgraphs} builds a compact object-centric 3D scene graph, combining geometric information with semantic descriptions and spatial relationships for scene perception and planning. ConceptFusion~\cite{jatavallabhula2023conceptfusion} further emphasizes open-set multimodal 3D mapping, allowing maps to be queried not only by language but also by images, audio, and 3D interactions. Together, these methods move scene understanding beyond closed-set labels toward reusable, multimodal, and queryable spatial representations.

For navigation specifically, VLFM~\cite{yokoyama2024vlfm} illustrates how such representations can support navigation by building occupancy maps from depth observations to identify frontiers, while using RGB observations and a pretrained VLM to generate a language-grounded value map for zero-shot semantic navigation. Multimodal Spatial Language Maps~\cite{huang2025multimodal} further extend vision-language maps by incorporating audio-vision-language information, enabling robots to ground text, image, or audio queries to spatial locations. 

Semantic maps are important for robot navigation in human environments because socially appropriate motion depends on more than obstacle geometry. Robots must understand where meaningful objects, functional areas, and interaction-relevant regions are located before deciding where to move, wait, yield, or explore.

\subsection{VLMs as Intermediate Representations}

A more deployment-oriented line of work treats VLM outputs less as final actions and more as intermediate planning representations that connect high-level reasoning to downstream planning and control. These representations may be explicit, such as schemas, programs, waypoint candidates, or attention maps, or implicit, such as latent embeddings. They can be used by local planners or lightweight policies.

This idea is related to a broader planning principle that intermediate navigation properties can be learned or inferred from prior planning experience and reused to guide hierarchical planning. For example, Stadler et al.~\cite{stadler2021online} learns high-level navigation properties online from previous planning results and use them to guide hierarchical multi-query planning. Within VLM-based navigation, NaviLLM~\cite{zheng2024towards} can be viewed as a schema-based interface, converting heterogeneous navigation tasks into a unified generation format. Yang et al.~\cite{yang2024octopus} uses a VLM as an intermediate representational layer that transforms multimodal task input into executable programs, bridging vision-language reasoning and embodied actions.

Other methods instantiate the intermediate representation more spatially. Chen et al.~\cite{chen2025affordances} prompts VLMs with visual annotations of navigable regions to produce waypoint or path candidates for continuous vision-language navigation. ViLAM~\cite{elnoor2025vi} distills VLM-guided social signals into attention maps that highlight regions relevant to socially compliant motion, as illustrated in Fig.~\ref{fig:taxonomy} (d). OLiVia-Nav~\cite{narasimhan2025olivia} provides a related latent-representation variant, where lightweight image and text encoders are trained using VLM-derived supervision for trajectory generation and selection.

These methods position VLMs less as direct decision makers and more as providers of reusable planning structures. Explicit spatial cues support local motion planning, while latent embeddings can encode a broader social and environmental context for efficient downstream policy learning or reuse. However, the effectiveness of these intermediate representations depends on whether they preserve the information needed for downstream safety, social compliance, and dynamic feasibility.

\subsection{VLMs as Action Interfaces}

A more tightly coupled paradigm connects vision-language reasoning directly to action interfaces. In this setting, the model does not merely recommend a goal, path, score, intermediate representation, or semantic map; instead, it produces action tokens, commands, or structured action representations conditioned on visual observations and language instructions, as illustrated in Fig.~\ref{fig:taxonomy} (e).

RT-2~\cite{zitkovich2023rt} is fine-tuned on web-scale vision-language tasks and robot trajectory data, representing robot actions as text tokens. This design allows semantic knowledge acquired from web-scale pretraining to influence robot action prediction. OpenVLA~\cite{kim2025openvla} and Octo~\cite{team2024octo} further generalize robot action learning by training generalist policies on heterogeneous robot datasets. Although these systems are mainly developed for manipulation rather than social navigation, they are important because they define a direct form of vision-language-to-action (VLA) coupling.

Navigation-oriented methods make this action interface more relevant to mobile robots. NaVid~\cite{zhang2024navid} uses a video-based VLM to encode language instructions together with the online history of RGB observations as spatio-temporal tokens, and then generates next-step navigation commands for closed-loop robot execution. 

Compared with intermediate-representation methods, this paradigm places the model output closer to executable action, ranging from action tokens and navigation commands to mid-level spatial-language actions. Code as Policies~\cite{liang2023code} provides an important precursor to hierarchical action interfaces by converting language instructions into executable programs that compose perception outputs, geometric computation, and robot control primitives. PaLM-E~\cite{driess2023palm} grounds a language model in visual and robotic sensor inputs, enabling it to generate high-level plans that are executed by low-level robot policies. NaVILA~\cite{cheng2024navila} uses a VLA model to generate mid-level spatial-language action commands and delegates real-time execution to a visual locomotion policy. Such interfaces aim to preserve the semantic and spatial reasoning benefits of VLMs while avoiding the need for large models to directly output low-level control commands at high frequency. 

Related navigation foundation policies integrate flexible goal specification directly into the control policy. ViNT~\cite{shah2023vint} learns visual navigation affordances from diverse navigation datasets and predicts both temporal distance to a visual subgoal and future actions. OmniVLA~\cite{hirose2025omnivla} continues this direction by training an end-to-end navigation policy conditioned on 2D poses, egocentric images, natural language, or their combinations.

This suggests a potential path toward SRN: a high-level VLM can ground abstract human intent in perceptual context, generate coarse action-level representations, and incorporate safety-aware reasoning, while a lower-level locomotion or navigation policy handles real-time motion execution~\cite{zhang2026walk}. However, directly coupling VLMs to action interfaces also raises challenges in latency, safety verification, distribution shift, and social compliance under dynamic human interactions.

\subsection{Efficient VLMs in Deployable SRN}

VLMs can provide strong contextual reasoning in SRN, but their computational cost and inference latency make them difficult to deploy on resource-constrained mobile robots. At the same time, lightweight VLMs are generally more suitable for low-latency inference, but they often suffer from weaker reasoning and decision-making in socially complex environments. Therefore, efficient VLMs are needed not only to reduce computational cost and latency but also to preserve socially aware reasoning and action accuracy under deployment constraints.

At the level of general VLM backbones, VILA~\cite{lin2024vila} studies how pre-training choices affect vision-language capability, efficiency, and deployability. For social navigation, Xiao and Yamasaki~\cite{xiao2026probing} show that prompting strategies can affect both efficiency and social-reasoning performance. 
E-SocialNav~\cite{xiao2026socialnav} further develops a lightweight model for socially compliant navigation under small-data settings. SocialNav-MoE~\cite{kawabata2025socialnavmoe} improves the efficiency--performance trade-off through a sparse mixture-of-experts (MoE) architecture and combines supervised, reinforcement-based, and MoE fine-tuning, including a semantic similarity reward for denser social-navigation feedback. 
Zhang et al.~\cite{zhang2026gcl} address performance limitations of lightweight VLMs through Group Competitive Learning (GCL), providing another training-level strategy for improving their reasoning capability.
MAction-SocialNav~\cite{11576553} demonstrates an efficient VLM design for deployable SRN. It uses a lightweight VLM to generate multiple plausible actions under action ambiguity, and further employs meta-cognitive prompting with a scoring signal and decision threshold to strengthen model’s reasoning capability. This enables socially aware action reasoning while preserving real-time efficiency, achieving 1.524 FPS.

These studies show that efficient VLM-based SRN requires advances across model architecture, adaptation strategy, and training design. General VLM backbones improve the deployability of vision-language reasoning, while prompt-level adaptation and lightweight fine-tuning reduce the cost of transferring VLMs to SRN tasks. At the same time, lightweight architectures and structured training strategies, such as sparse expert models, reinforcement-based fine-tuning, and competitive learning, help preserve socially aware reasoning and action accuracy under limited data and hardware constraints. Overall, these methods move VLM-based SRN closer to practical, real-time, and socially aware robot deployment. However, how to jointly maintain low latency, accuracy, safety, and robustness under real-world distribution shifts remains an open challenge.

\section{Datasets and Evaluation Platforms}
\label{sec:datasets}
\subsection{Datasets}

\begin{table*}[t]
\centering
\footnotesize
\setlength{\tabcolsep}{2pt}
\renewcommand{\arraystretch}{1.08}
\caption{Datasets relevant to VLM-based SRN}
\label{tab:dataset_comparison}
\begin{tabular}{@{}>{\raggedright\arraybackslash}p{0.15\textwidth}
                >{\raggedright\arraybackslash}p{0.15\textwidth}
                >{\raggedright\arraybackslash}p{0.17\textwidth}
                >{\raggedright\arraybackslash}p{0.17\textwidth}
                >{\raggedright\arraybackslash}p{0.15\textwidth}
                >{\raggedright\arraybackslash}p{0.17\textwidth}@{}}
\toprule
Dataset & Viewpoint & Temporal Horizon & Language Instruction & Social Context & Output Annotation \\
\midrule
SDD~\cite{robicquet2016learning} & Bird's-eye view & Long-horizon & Unavailable & Explicit & Trajectory \\
EgoMotion~\cite{park2016egocentric} & Human egocentric & Short-horizon & Unavailable & Implicit & Trajectory \\
WILDTRACK~\cite{chavdarova2018wildtrack} & Third-person & Short-horizon & Unavailable & Implicit & Trajectory \\
JRDB~\cite{martin2019jrdb} & Robot egocentric & Short-horizon & Unavailable & Implicit & Trajectory \\
THOR~\cite{rudenko2020thor} & Bird's-eye view & Long-horizon & Unavailable & Explicit & Trajectory \\
SocNav2~\cite{bachiller2022graph} & Bird's-eye view & Long-horizon & Unavailable & Explicit & Trajectory \\
DynaBARN~\cite{nair2022dynabarn} & Robot egocentric & Long-horizon & Available & None & Action \\
SCAND~\cite{karnan2022socially} & Robot egocentric & Long-horizon & Available & Explicit & Action \\
MuSoHu~\cite{nguyen2023toward} & Human egocentric & Long-horizon & Available & Explicit & Trajectory \\
SG-LSTM~\cite{bhaskara2023sg} & Robot egocentric & Short-horizon & Unavailable & Explicit & Trajectory \\
HuRoN~\cite{hirose2023sacson} & Robot egocentric & Short-horizon & Available & Explicit & Action \\
SNEI~\cite{payandeh2025social} & Robot egocentric & Short-horizon & Available & Explicit & Action \\
MUSON~\cite{liu2025muson} & Human egocentric & Short-horizon & Available & Explicit & Action \\
GazeNav~\cite{yu2026learning} & Human egocentric & Short-horizon & Available & Explicit &  Trajectory \\
\bottomrule
\end{tabular}
\end{table*}

Early trajectory datasets mainly provide trajectory annotations and scene context for understanding pedestrian motion. SDD~\cite{robicquet2016learning} offers long-horizon bird's-eye view trajectories in real-world campus scenes, with semantic scene masks and explicit target--target or target--space interactions, making it useful for studying social motion patterns but not robot-centered decision making. EgoMotion~\cite{park2016egocentric} uses the human egocentric visual experience to predict future self-motion; it is useful for studying the navigational value of first-person vision, although its social cues remain implicit. WILDTRACK~\cite{chavdarova2018wildtrack} and JRDB~\cite{martin2019jrdb} provide dense pedestrian perception from third-person and robot egocentric viewpoints, respectively, but their annotations mainly support detection and tracking rather than explicit social reasoning or policy learning.

Several datasets move closer to social navigation by adding structured scene representations, social labels, or task annotations. THOR~\cite{rudenko2020thor} records human trajectories in controlled indoor human-robot scenarios with task roles, group information, and gaze/head orientation, making it useful for modeling socially situated motion intent. SocNav2~\cite{bachiller2022graph} directly targets human-aware robot navigation by representing dynamic scenes as graphs and supervising robot behavior with human social-acceptability scores. DynaBARN~\cite{nair2022dynabarn} instead focuses on dynamic obstacle navigation: it provides moving obstacles following waypoint-based trajectories and teleoperated demonstrations, but contains little explicit social reasoning.

More recent robot-navigation datasets provide richer demonstrations, egocentric observations, or social-trajectory supervision. SCAND~\cite{karnan2022socially} records robot egocentric sensor streams and joystick demonstrations in real campus environments, with social interaction labels that make it suitable for imitation learning of socially compliant navigation. MuSoHu~\cite{nguyen2023toward} captures human egocentric navigation behavior in social environments, providing human trajectories and waypoint-level navigation cues for learning from human demonstrations. SG-LSTM~\cite{bhaskara2023sg} focuses on group-aware trajectory prediction from a robot-centered view, using graph representations to bridge perception and social grouping.

The latest datasets connect perception, social reasoning, and action more directly. HuRoN~\cite{hirose2023sacson} introduces reward or policy-level supervision for navigation in human environments, making it closer to robot decision learning than pure trajectory prediction. SNEI~\cite{payandeh2025social} and MUSON~\cite{liu2025muson} add rationale style supervision, which is especially relevant for VLM-based systems because it can align visual observations with interpretable social judgments. GazeNav~\cite{yu2026learning} further incorporates attention cues from a human egocentric viewpoint, providing an explicit bridge between visual attention and navigation intent.

Existing datasets cover complementary parts of the VLM-based SRN pipeline. Perception-oriented datasets provide visual grounding and human trajectories; social navigation datasets add norms, interaction labels, scores, or graph structures; and recent rationale- or gaze-based datasets begin to expose the kind of interpretable intermediate reasoning that VLMs are well suited to exploit. However, to the best of our knowledge, no single dataset fully combines egocentric robot vision, language instruction or rationale, explicit social reasoning annotations, bridging representations, and executable robot actions, leaving a clear gap for future dataset construction.

\subsection{Evaluation Platforms}
\label{sec:Evaluation-Platforms}

While existing datasets primarily support evaluating whether a model can perceive and interpret human-centered scenes, SRN ultimately requires translating such understanding into navigation actions. Evaluating this capability requires interactive evaluation platforms that support closed-loop human-robot interaction. Moreover, the persistent simulation-to-reality gap remains a major challenge in robot navigation~\cite{gervet2023navigating}, making it essential to assess whether high-level VLM understanding can be transformed into safe, efficient, and socially appropriate navigation behaviors in realistic environments.

Because visual observations are central to VLM-based SRN, we organize representative evaluation platforms according to four behavior-centric dimensions: the evaluation target, the social factors present in the environment, the evaluation metrics, and the application scenarios, as summarized in Table~\ref{tab:benchmark_comparison}.

\begin{table*}[t]
\centering
\footnotesize
\setlength{\tabcolsep}{2pt}
\renewcommand{\arraystretch}{1.08}
\caption{Evaluation platforms for SRN}
\label{tab:benchmark_comparison}
\begin{tabular}{@{}>{\raggedright\arraybackslash}m{0.15\textwidth}
                >{\centering\arraybackslash}m{0.10\textwidth}
                >{\raggedright\arraybackslash}m{0.14\textwidth}
                >{\raggedright\arraybackslash}m{0.12\textwidth}
                >{\raggedright\arraybackslash}m{0.22\textwidth}
                >{\raggedright\arraybackslash}m{0.20\textwidth}@{}}
\toprule
Platform & Type & Evaluation Target & Social Factors & Metrics & Application Scenario \\
\midrule

gym-collision-avoidance~\cite{everett2018motion}
& Simulator
& Obstacle avoidance
& Agent
& Collision cases, stuck cases, average extra time to goal
& Crowd collision avoidance \\ \midrule

iGibson 2.0~\cite{li2021igibson}
& Simulator
& Object-centric household manipulation
& Object state
& Success rate
& Indoor service navigation \\ \midrule

Intelligent Human Avatar~\cite{favier2022intelligent}
& Simulator
& Human-aware navigation
& Human reaction
& Human time to goal
& Doorway conflict navigation \\ \midrule

SocialGym~\cite{holtz2022socialgym}
& Simulator
& Social compliance
& Human
& Successful trials, force, blame, time to goal
& Dynamic social navigation \\ \midrule

SEAN 2.0~\cite{tsoi2022sean}
& Simulator
& Human-aware navigation
& Pedestrian, social situation
& Path efficiency, time not moving, completed tasks, collision rate
& Context-aware social navigation \\ \midrule 

HuNavSim~\cite{perez2023hunavsim}
& Simulator
& Human-aware navigation
& Human reaction
& Social compliance metrics
& Reactive human-aware navigation \\ \midrule

Habitat 3.0~\cite{puig2023habitat}
& Simulator
& Human-aware navigation
& Humanoid, obstacle
& Following rate, collision rate
& Humanoid-following social navigation \\ \midrule

DynaBARN~\cite{nair2022dynabarn}
& Benchmark
& Obstacle avoidance
& Obstacle
& Success rate
& Dynamic obstacle avoidance \\ \midrule 

SocNavBench~\cite{biswas2022socnavbench}
& Benchmark
& Human-aware navigation
& Pedestrian
& Path quality, motion quality, pedestrian disruption, meta statistics
& Pedestrian-aware social navigation \\ \midrule

Arena-Bench~\cite{kastner2022arena}
& Benchmark
& Obstacle avoidance
& Pedestrian
& Collision rate, success rate
& Obstacle avoidance in simulation and real-world environments \\ \midrule

Social Nav. Protocol~\cite{pirk2022protocol}
& Benchmark
& Social compliance
& Social norm
& Human-rated social compliance
& Real-world social norm navigation \\ \midrule

Human Empowerment Evaluation~\cite{baddam2025search}
& Benchmark
& Social compliance
& Human-robot interaction
& Social compliance
& Human-empowerment-aware crowd navigation \\ \midrule

SocialNav-SUB~\cite{munje2025socialnav}
& Benchmark
& Social navigation scene understanding
& Human
& Probability of agreement
& Evaluation of VLMs for SRN \\

\bottomrule
\end{tabular}
\end{table*}

Early learning-based navigation simulators such as gym-collision-avoidance~\cite{everett2018motion} mainly evaluate whether an agent can avoid other moving agents and reach the goal efficiently. They are useful for testing low-level collision avoidance, but they do not evaluate whether the robot understands social context or whether language-based reasoning improves navigation.
iGibson 2.0~\cite{li2021igibson} provides rich visual and semantic grounding for household object-centric tasks. Although it is not specifically designed for SRN, it offers realistic embodied environments that are valuable for studying visually grounded decision making.
Human-avatar and human-reaction simulators such as Intelligent Human Avatar~\cite{favier2022intelligent} and HuNavSim~\cite{perez2023hunavsim} introduce interactive human behaviors and reactive agents. They are useful for testing whether a navigation policy can handle doorway conflicts, blocking, curiosity, fear, or other socially relevant human responses.
SocialGym~\cite{holtz2022socialgym} and related extensions such as SocNavGym~\cite{kapoor2023socnavgym} and SOCIALGYM 2.0~\cite{sprague2023socialgym} explicitly target social navigation. They introduce social action selection, learned social rewards, group interactions, human-object interactions, and social mini-games such as doorways, hallways, intersections, and roundabouts. These platforms are well suited for evaluating the behavioral consequences of VLM-based SRN policies.
SEAN 2.0~\cite{tsoi2022sean} and Habitat 3.0~\cite{puig2023habitat} are especially relevant to visually grounded social navigation because they provide embodied simulated worlds with humans or humanoids, robot sensors, and social navigation tasks. Still, their standard evaluations focus on task completion, following distance, collisions, or proxemics, not on natural-language instruction following or explanation-based navigation.
SocialNav-SUB~\cite{munje2025socialnav} provides a complementary benchmark perspective, showing that the evaluated VLMs still struggle with spatial, spatiotemporal and social scene understanding.

Beyond simulation platforms, several benchmarks provide standardized protocols for evaluating navigation performance and social behavior. DynaBARN~\cite{nair2022dynabarn}, SocNavBench~\cite{biswas2022socnavbench}, and Arena-Bench~\cite{kastner2022arena} focus on evaluating navigation performance in dynamic environments. DynaBARN and Arena-Bench primarily assess obstacle avoidance under moving obstacles or pedestrians, while SocNavBench extends evaluation to human-aware navigation through metrics such as trajectory quality, pedestrian disruption, and safety. However, these benchmarks still focus on navigation outcomes rather than language instructions, social explanations, or rationale faithfulness.
The Social Navigation Protocol~\cite{pirk2022protocol} is one of the few real-world benchmarks, using human ratings to evaluate whether robot behavior satisfies social expectations. This makes it valuable for validating socially acceptable navigation behavior in real environments.
Human Empowerment Evaluation~\cite{baddam2025search} adds an important perspective by measuring whether the robot preserves human agency during crowd navigation. This provides a useful social-compliance metric for VLM-based SRN behavior evaluation, although it remains a trajectory-level assessment and does not evaluate VLM-generated explanations or reasoning processes.

\section{Challenges and Future Directions}
\label{sec:challenges_future}

VLM-based SRN must connect high-level VLM reasoning with low-level planning and control. Existing work suggests that a central difficulty lies not only in improving perception, planning or control in isolation, but also in aligning the full pipeline: what the robot observes, which social meaning it infers, how this meaning is represented for planning, and how the resulting behavior is evaluated.

\subsection{High-Quality SRN Datasets}
One central bottleneck is the lack of datasets that jointly align perception inputs, social annotations, planning representations, action labels, and human feedback. For example, SCAND~\cite{karnan2022socially} covers diverse social navigation interactions but was collected in a single city and may reflect region-specific navigation patterns. Recent VLM-based SRN benchmarks also remain constrained by limited coverage of models and scenarios, as well as reliance on a small set of source datasets~\cite{munje2025socialnav}. More broadly, controlled datasets may not capture the diversity of real-world SRN conditions such as cultural norms, local traffic conventions, and rare human-robot interactions~\cite{xiao2026probing}.

Therefore, future datasets should move beyond scale alone. They should provide structured supervision for social cues, spatial constraints, and causal rationales~\cite{liu2025muson}, long-horizon annotated videos for embodied navigation~\cite{zhang2024navid}, and broader distributions of robot platforms, skills, and action trajectories so that VLM pretraining can transfer to new physical behaviors rather than only new semantic concepts~\cite{zitkovich2023rt}. A promising direction for expanding SRN datasets is to use simulation and synthetic generation as a complement to real-world human-robot interaction data~\cite{francis2025principles}.

\subsection{Robust Social and Spatial Reasoning}

VLMs offer a promising interface for interpreting social scenes, but their reasoning can remain fragile in dynamic navigation settings. Prior work reports inconsistent VLM inferences~\cite{zhang2023grounding}, loss of semantic information when egocentric visual observations are compressed into text memory~\cite{zhang2025mem2ego}, and failures when the model has access only to the current frame and misses social context outside the camera view~\cite{fang2026obstacles}. Language grounding is also incomplete because landmark-based navigation can ignore verbs and nuanced commands such as speed, manner, or social constraints~\cite{shah2023lm}, while vision-and-language navigation requires a tight coupling between local perception and long-term decision making~\cite{fried2018speaker}.

Future work should strengthen temporal memory, spatial grounding, and open-world social understanding. Static semantic maps and vision-language maps need to handle dynamic humans and moving objects~\cite{chen2023open,huang2023visual,xia2018gibson}. Dynamic open-vocabulary mapping should further incorporate human-object interactions and human activities, which remain underrepresented in many current systems~\cite{jiang2025dualmap}. Because social norms are often ambiguous and vary across individuals, cultures, and contexts~\cite{kawabata2025socialnavmoe}, VLM-based SRN systems should represent uncertainty rather than forcing a single deterministic interpretation or action.

\subsection{Reasoning to Control Grounding}

The key challenge is not only whether a VLM can describe a scene, but whether its reasoning can be grounded in executable motion. Several works point to this missing bridge. VLMs can serve as interfaces to controllers that were not trained with language-labeled navigation data~\cite{shah2023lm}, but the connection between high-level model outputs and robot control is still underdeveloped. Graph-based social reasoning, for example, still requires mechanisms for linking GNN outputs to navigation control, possibly as constraints for model predictive control~\cite{bachiller2022graph}. Similarly, hierarchical 3D scene graphs may support efficient planning, but their use as representations for downstream control remains underexplored~\cite{gu2024conceptgraphs}.

Future VLM-based SRN systems should make the bridging representation explicit. Possible forms include social cost maps, reward functions, waypoints, constraints, attention maps or gaze cues, and interaction graphs. However, these bridges must be carefully validated. VLM-based reward models can be misspecified when prompts underspecify human intent or when the VLM generalizes poorly to unseen social contexts and environments~\cite{rocamonde2023vision}. Hand-designed rewards also require delicate tuning between path following and human avoidance, motivating reward learning, inverse reinforcement learning, and preference learning~\cite{han2025dr}. Human-robot trajectory datasets such as SCAND can support learning socially compliant cost functions~\cite{karnan2022socially}, but future work should evaluate whether such learned bridges improve closed-loop socially compliant navigation behavior.

\subsection{Real-Time Embodied Control}

VLM-based SRN must operate under real robot constraints. Dense-crowd navigation may require replanning at around 10 Hz to avoid accumulating tracking errors~\cite{trautman2015robot}. At the same time, large VLM-based navigation models and VLA policies introduce substantial inference cost, which can become a bottleneck for high-frequency control~\cite{shah2023vint,zhang2024navid,zitkovich2023rt}. Many current evaluations also simplify the action space, for example, by using discrete actions, which limits realism compared with continuous control in embodied environments~\cite{xiao2026probing}.

Future systems should combine VLM reasoning with lightweight and reliable control modules. Promising directions include quantization, distillation, action chunking, and hybrid architectures that reserve VLMs for social reasoning while using specialized planners or controllers for fast execution~\cite{zhang2024navid,zitkovich2023rt,munje2025socialnav}. Systems should support richer action spaces, diverse robot morphologies, and multimodal sensors~\cite{shah2023vint,sprague2023socialgym}. Since collision avoidance remains difficult for language-conditioned policies~\cite{hirose2024lelan}, the bridge from VLM output to low-level planning and control should be tested under continuous, closed-loop, human-reactive navigation.

\subsection{Continual Adaptation}
Continual and online adaptation is another important direction for VLM-based SRN. Existing work suggests that static offline-trained models may be insufficient for dynamic social environments. MUSON~\cite{liu2025muson} motivates reinforcement learning and continual adaptation as a way to align reasoning with sequential decision making under real-world constraints, while SELFI~\cite{hirose2024selfi} emphasizes human-in-the-loop online learning based on human evaluations. Other works further highlight online prompt adaptation~\cite{xiao2026probing}, online refinement of social priors~\cite{elnoor2025vi}, and continual feedback for dynamically computing proxemic constraints~\cite{narayanan2020proxemo}. These directions suggest that future SRN systems should not only learn from fixed datasets, but also adapt continuously from human feedback, environmental changes, and deployment-time interaction.

\subsection{Evaluation Beyond Success and Collision}

Current benchmarks often evaluate whether the robot reaches the goal safely, but VLM-based SRN also requires evaluation of social reasoning, reasoning-action consistency, and human acceptance. Existing metrics can expose trade-offs between task success and social comfort, but may fail to capture social behaviors~\cite{gong2025cognition}. Social compliance metrics themselves also require validation through user studies, since automatic metrics may not fully reflect human perception~\cite{baddam2025search}. Moreover, SRN evaluation is complicated by local norms, human preferences, and ambiguous ground truth~\cite{karnan2022socially,kawabata2025socialnavmoe}.

Evaluating the full VLM-based SRN loop requires richer conflict scenarios and metrics~\cite{favier2022intelligent}, broader VLM and dataset coverage~\cite{munje2025socialnav}, and human-in-the-loop feedback for online adaptation~\cite{hirose2024selfi}. Rather than only reporting success rate, collision rate, or time-to-goal, future evaluations should include proxemics, comfort, social compliance, human preference, rationale--action consistency, and instruction-following correctness. Such evaluations should assess not only final trajectories, but also intermediate reasoning outputs and planner-facing representations.

\subsection{Generalization and Sim-to-Real}

VLM-based SRN must generalize not only across physical environments, but also across cultures, robot embodiments, and human-robot interaction styles. Laboratory studies cannot fully reproduce real pedestrian behavior~\cite{mavrogiannis2019effects}, and findings from one region, robot embodiment, or deployment site may not transfer directly to another~\cite{senft2020would,fujioka2024need}. Culture makes this transfer harder because social norms, personal space, passing preferences, and polite behaviors are context-dependent rather than universal~\cite{kruse2013human,rios2015proxemics,moller2021survey,francis2025principles}.

Fixed personal-space costs and hand-coded social rules may be insufficient. VLM-based SRN systems need to adapt to local norms and human preferences, and evaluation should reflect culture-dependent comfort, perceived safety, social compliance, and acceptance~\cite{wang2022metrics,mavrogiannis2023core,singamaneni2024survey,charalampous2017recent}. Simulation adds another layer of uncertainty, with gaps in human behavior, robot dynamics, sensing, and pedestrian-robot interaction~\cite{xu2023safecrowdnav,luo2025goal}. These gaps are especially critical because pedestrians may react to, negotiate with, or intentionally block a robot in ways absent from curated datasets~\cite{liu2022intention,han2025dr,mirsky2024recognition}.

Future simulators should model more realistic human reactions and support sim-to-real transfer~\cite{karnan2022socially,xu2023safecrowdnav}. Real-world evaluation should cover diverse pedestrians, robot embodiments, social norms, and long-horizon interactions, helping VLM-based SRN move from visually plausible social reasoning toward reliable and socially acceptable navigation in shared human spaces.

\section{Conclusion}
\label{sec:conclusion}

This survey examined VLM-based SRN through the lens of a key literature gap. VLM-centric approaches provide rich semantic understanding and commonsense reasoning but often struggle to satisfy the real-time, safety, and reliability requirements of robotic deployment. In contrast, deployment-ready navigation systems remain largely limited to geometric reasoning and lack rich semantic awareness of human-centered environments. To address this gap, we reviewed existing efforts that integrate VLMs into SRN pipelines and analyzed how different interface designs connect high-level VLM reasoning to low-level planning and control. The reviewed evidence suggests that hierarchical hybrid architectures are currently among the most practical directions for combining semantic reasoning with reliable navigation performance.

Despite this progress, significant challenges remain before VLM-based SRN can achieve widespread real-world deployment. A key challenge is ensuring that high-level VLM reasoning can be translated into real-time, safe, and executable navigation actions. This requires addressing latency bottlenecks, improving spatial grounding and reasoning-to-action consistency, and developing robust safety mechanisms, including formal verification and reliable fallback strategies for VLM-in-the-loop systems. At the same time, fragmented datasets, limited benchmarking standards, temporal mismatch between reasoning and execution, cultural variability, continual adaptation, and long-horizon decision making continue to hinder systematic progress and fair evaluation across approaches.

Looking forward, the central challenge of the field is not whether semantic reasoning can enhance SRN, but how such reasoning can be reliably, safely, and efficiently integrated into embodied navigation systems. Future advances will likely depend on tighter coupling between semantic reasoning, environmental grounding, and physical interaction, ultimately enabling socially compliant navigation systems that are semantically informed, safe, and deployable in the real-world environments.

\bmhead{Acknowledgements}

This work was supported in part by JST Moonshot Goal 3 under Grant JPMJMS263E-12 and the NVIDIA Academic Grant Program.

\bibliography{ref-survey}

\end{document}